\documentclass[letterpaper]{article} 
\usepackage{aaai2027}
\usepackage[hyphens]{url}  
\usepackage{graphicx} 
\usepackage{natbib}  
\usepackage{caption} 
\usepackage{amsmath}
\usepackage{amssymb}
\usepackage{amsfonts}
\usepackage{booktabs}
\usepackage{tabularx}
\usepackage{array}
\usepackage{colortbl}
\usepackage{microtype}

\newcolumntype{Y}{>{\raggedleft\arraybackslash}X}
\newcolumntype{L}{>{\raggedright\arraybackslash}X}

\title{Region-Constrained Group Relative Policy Optimization\\for Flow-Based Image Editing}

\author{
    Zhuohan Ouyang\equalcontrib\textsuperscript{\rm 1},
    Zhe Qian\equalcontrib\textsuperscript{\rm 2,\rm 3},
    Wenhuo Cui\textsuperscript{\rm 1},
    Chaoqun Wang\corresponding\textsuperscript{\rm 1}
}
\affiliations{
    \textsuperscript{\rm 1}South China Normal University\\
    \textsuperscript{\rm 2}South China Agricultural University\\
    \textsuperscript{\rm 3}Monash University
}

\begin{document}

\maketitle

\begin{abstract}
Instruction-guided image editing requires balancing target modification with non-target preservation. Recently, flow-based models have emerged as a strong backbone for instruction-guided image editing, thanks to their high fidelity and efficient deterministic ODE sampling. Building on this foundation, GRPO-based reward-driven post-training has been explored to directly optimize editing-specific rewards, improving instruction following and editing consistency. However, existing methods often suffer from noisy credit assignment: global exploration also perturbs non-target regions, inflating within-group reward variance and yielding noisy GRPO advantages.
To address this, we propose RC-GRPO-Editing, a region-constrained GRPO post-training framework for flow-based image editing under deterministic ODE sampling. It suppresses background-induced nuisance variance to enable clean localized credit assignment, improving editing region instruction adherence while preserving non-target content.
Concretely, we localize exploration via region-decoupled initial noise perturbations, which reduce background-induced reward variance and stabilize GRPO advantages. We further introduce an attention concentration reward that aligns cross-attention with the intended editing region throughout the rollout, reducing unintended changes in non-target regions.
Experiments on CompBench, GEdit, and ImgEdit show consistent gains in editing region instruction adherence and non-target preservation.
\end{abstract}

\section{Introduction}

Instruction-guided image editing aims to modify a given image according to a target instruction, for example by adding, removing, or altering visual content and attributes, while preserving non-target content and the overall scene structure~\cite{brooks2023instructpix2pix,meng2022sdedit,hertz2023prompttoprompt}.
Recently, flow-based models~\cite{lipman2022flow,liu2022flow,albergo2023building,song2021scorebased} have been increasingly adopted as a strong backbone for instruction-guided image editing~\cite{labs2025flux}, thanks to their strong realism and efficient deterministic ODE sampling.

In parallel, reward-driven post-training (e.g., GRPO-style optimization) has emerged as a practical way to improve instruction following and edit locality beyond supervised fine-tuning, by directly optimizing task-level rewards~\cite{shao2024deepseekmath,liu2025omnirefiner,tan2026talk2move,schulman2017proximal}. Nevertheless, extending GRPO-style post-training to these deterministic ODE editors remains non-trivial.

\begin{figure}[t]
    \centering
    \includegraphics[width=\columnwidth,trim=12 54 74 86,clip,keepaspectratio]{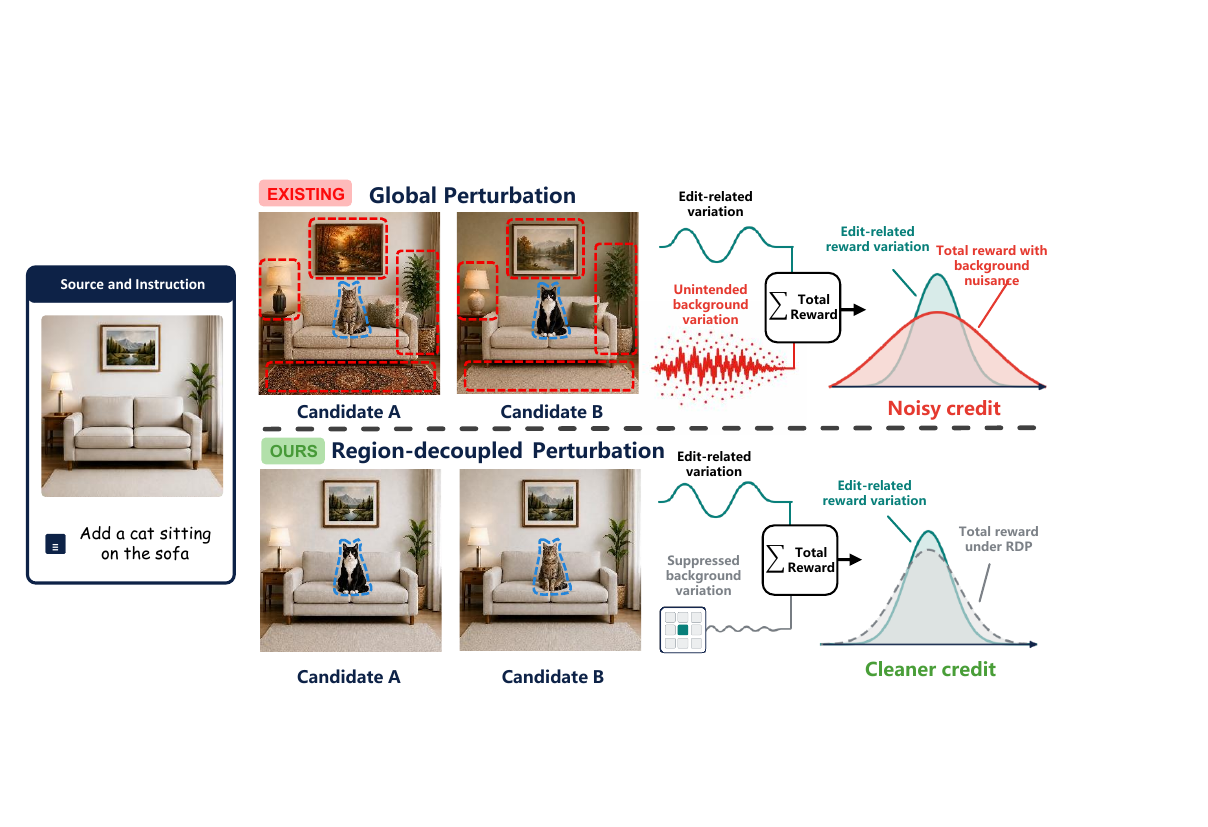}
    \vspace{-6mm}
    \caption{\textbf{Motivation for region-decoupled perturbation.}
    Global perturbation changes both the editing region and the background, so variation in the preservation reward introduces nuisance variance into the total reward and produces noisy credit assignment.
    RDP suppresses unintended background variation, yielding more reliable credit assignment.}
    \label{fig:dynamic_relation}
    \vspace{-0.8mm}
\end{figure}

To enable GRPO-style post-training for deterministic ODE samplers, prior work injects exploration either by an ODE-to-SDE conversion that yields an equivalent SDE matching the original model's marginal distribution at all timesteps~\cite{liu2025flow,song2021scorebased}, or by a mixed ODE-SDE strategy with a sliding-window mechanism~\cite{li2025mixgrpo}.
Alternatively, Neighbor GRPO~\cite{he2025neighbor} perturbs the initial noise to construct a neighborhood of candidate ODE trajectories and optimizes a softmax distance-based surrogate leaping policy, preserving deterministic ODE sampling and compatibility with high-order solvers~\cite{lu2022dpm}.
However, in image editing, these exploration mechanisms exhibit a key mismatch: global initial noise perturbations also randomize non-target regions, inflating within-group reward variance and lowering the SNR of GRPO advantages~\cite{schulman2017proximal}. This yields noisy credit assignment for the target edit and often manifests as a trade-off between editing region instruction adherence and non-target preservation (see Figure~\ref{fig:dynamic_relation}). Recent editing-oriented GRPO methods address this at the credit-assignment stage, decomposing the advantage across regions~\cite{xu2026editgrpo,long2026spatialreward,yang2026spatialflow} or denoising steps~\cite{savani2026stepwise}. We instead act at the exploration stage, before the advantage is formed.

In this work, we propose \textbf{RC-GRPO-Editing}, a region-constrained GRPO post-training framework for flow-based image editing under deterministic ODE sampling, which improves localized credit assignment and further strengthens editing region instruction adherence while maintaining non-target preservation. The framework combines region-constrained exploration and an attention-based intrinsic reward to align GRPO updates with the editing region.

To reduce background-induced nuisance variance and improve localized credit assignment,
we design \textbf{Region-Decoupled Perturbation (RDP)} to localize GRPO exploration to the intended editing region.
Concretely, during post-training we construct a group of neighboring initial conditions by perturbing the initial noise only within the target region while suppressing perturbations elsewhere, so that candidate rollouts remain diverse where edits are expected but exhibit minimal unintended variation in non-target areas.
To provide GRPO with a clean, locality-aware reward signal,
we design \textbf{Attention Concentration Density (ACD)} as an intrinsic reward based on text-to-image cross-attention concentration along the rollout.
Specifically, ACD quantifies the degree to which text-to-image cross-attention is concentrated within the intended editing region along the trajectory. This encourages policy updates to prioritize edits within the intended editing region and reduces gradient noise from non-target regions.

Our contributions are summarized as follows:
\begin{itemize}
    \item We propose RC-GRPO-Editing, a region-constrained GRPO post-training framework for deterministic ODE image editing that improves localized credit assignment to strengthen editing region adherence and preserve non-target content.

    \item We introduce two components: RDP, which perturbs the initial noise only in the editing region to reduce background-induced variance; and ACD, an intrinsic reward that encourages cross-attention to stay in the editing region and provides an interpretable diagnostic.

    \item Experiments and ablations show consistent improvements on foreground instruction adherence and background preservation, validate the complementary roles of RDP and ACD, and further demonstrate gains across two different editing backbones.
\end{itemize}

\section{Related Work}
\label{sec:related_work}

\subsection{Instruction-Guided Image Editing}
Instruction-guided editors range from supervised models such as InstructPix2Pix~\cite{brooks2023instructpix2pix} and MagicBrush~\cite{zhang2023magicbrush} to recent unified systems such as Step1X-Edit~\cite{liu2025step1x}, Qwen-Image-Edit~\cite{wu2025qwenimage}, and GoT~\cite{fang2025got}. Training-free methods instead steer latent trajectories, embeddings, or cross-attention~\cite{meng2022sdedit,hertz2023prompttoprompt,mokady2023null,avrahami2022blended,couairon2023diffedit}. Spatial conditioning and attention modulation further reduce unintended changes~\cite{zhang2023adding,guo2024focus}, but reliably confining an edit remains difficult. We use cross-attention as a training signal and focus on reward-driven post-training of flow-based editors with deterministic ODE sampling~\cite{lipman2022flow,liu2022flow,albergo2023building}.

\subsection{Reward-Driven Post-Training and GRPO-Style Optimization}
Reward-driven fine-tuning and preference optimization improve generative alignment~\cite{black2024training,wallace2024diffusion,yang2024using}. For flow models, Flow-GRPO~\cite{liu2025flow} and MixGRPO~\cite{li2025mixgrpo} introduce SDE-based exploration, while Neighbor GRPO~\cite{he2025neighbor} perturbs the initial noise to form deterministic ODE neighborhoods. Editing-oriented methods also apply RL to localized refinement and geometric transformations~\cite{liu2025omnirefiner,tan2026talk2move}.

Recent methods improve credit assignment by decomposing group-normalized advantages across image regions~\cite{xu2026editgrpo,long2026spatialreward,yang2026spatialflow} or denoising steps~\cite{savani2026stepwise}. They retain global exploration and modify credit after rollouts are sampled. In contrast, RDP constrains exploration before advantage estimation, while ACD penalizes cross-region attention coupling. Our approach is therefore complementary to region-aware credit assignment.

\section{GRPO for Deterministic ODE Image Editing}

\subsection{Preliminaries}

\subsubsection{Flow-Based ODE Formulation.}
Given a source image $I_{\mathrm{src}}$ and instruction $p$, let $c=(I_{\mathrm{src}},p)$ denote the editing conditions.
A flow-based model defines a deterministic generative ODE~\cite{liu2022flow,song2021scorebased}:
\begin{equation}
\frac{d x_t}{dt} = v_\theta(x_t, t, c),\qquad t\in[0,1].
\label{eq:ode}
\end{equation}

Integrating Eq.~\eqref{eq:ode} from $x_1=\epsilon\sim\mathcal{N}(0,I)$ to $x_0$ with a fixed ODE solver induces the deterministic map~\cite{chen2018neural,grathwohl2019ffjord}:
\begin{equation}
x_0 = \Phi_\theta(\epsilon, c), \qquad \epsilon \sim \mathcal{N}(0,I),
\label{eq:det_map_main}
\end{equation}
where $\Phi_\theta$ maps the initial noise $\epsilon$ to the output under conditions $c$.

\subsubsection{GRPO via Initial Noise Neighborhoods.}
Deterministic ODE sampling lacks the stochastic exploration required by policy gradient methods~\cite{schulman2017proximal}. Neighbor GRPO~\cite{he2025neighbor} instead perturbs the initial noise to construct a local solution neighborhood while keeping each rollout deterministic. Given terminal rewards $\{r_i\}_{i=1}^G$ for $G$ rollouts, GRPO defines the group-normalized advantage~\cite{shao2024deepseekmath}:
\begin{equation}
A_i=\frac{r_i-\mathrm{mean}(\{r_1,\ldots,r_G\})}{\mathrm{std}(\{r_1,\ldots,r_G\})}.
\label{eq:grpo_adv}
\end{equation}

\begin{figure*}[t]
    \centering
    \includegraphics[width=0.98\textwidth,keepaspectratio]{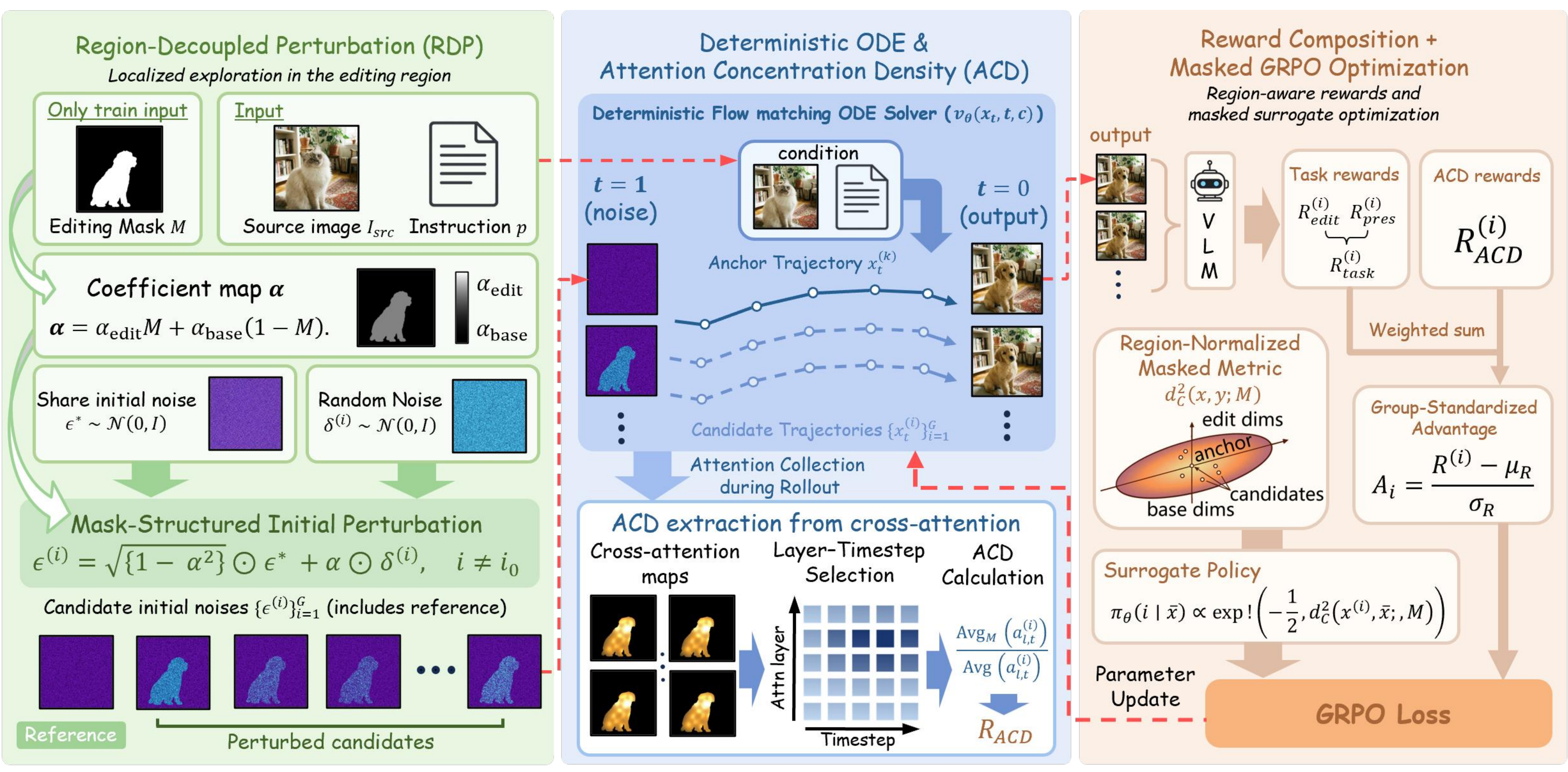}
    \vspace{-0.6mm}
    \caption{\textbf{Method overview.}
    RDP constructs a mask-structured initial noise neighborhood at $t{=}1$ to localize exploration to the editing region.
    Deterministic ODE rollouts from $t{=}1\rightarrow 0$ provide candidate trajectories, and ACD computes an intrinsic reward from cross-attention concentration within the mask.
    GRPO combines VLM task rewards and ACD to update the model using a mask-aware surrogate policy over candidates.}
    \label{fig2}
\end{figure*}

GRPO uses these advantages in the standard clipped policy-ratio objective. Given a base initial noise $\epsilon^\ast\sim\mathcal{N}(0,I)$, Neighbor GRPO constructs $G$ correlated initial conditions:
\begin{equation}
\epsilon^{(i)}=\sqrt{1-\sigma^2}\,\epsilon^\ast+\sigma\,\delta^{(i)},
\label{eq:ngrpo_init}
\end{equation}
where $\sigma\in(0,1)$ controls the perturbation strength and $\delta^{(i)}\stackrel{\text{i.i.d.}}{\sim}\mathcal{N}(0,I)$. Because each rollout is deterministic after $t=1$, Neighbor GRPO defines a training-only surrogate leaping policy over candidates. At timestep $t$, let $k$ be the sampled anchor and $x_t^{(\theta)}$ its trajectory re-integrated under $\theta$. The policy is a softmax over negative squared distances:
\begin{equation}
\pi_\theta\!\left(i \mid t,k\right)=
\frac{\exp\!\left(-\|x_t^{(i)}-x_t^{(\theta)}\|_2^2/2\right)}
{\sum_{j=1}^{G}\exp\!\left(-\|x_t^{(j)}-x_t^{(\theta)}\|_2^2/2\right)}.
\label{eq:leap_policy}
\end{equation}

\subsection{Motivation: Why Global Perturbation Hurts Credit Assignment}
\label{sec:motivation}

In Neighbor GRPO, intra-group diversity arises solely from perturbations of the initial noise.
For image editing, however, global perturbations randomize both the editing region and the background,
introducing background-induced \emph{nuisance variance}.
This variance is unrelated to localized instruction following and degrades the signal-to-noise ratio (SNR) of the group-standardized advantages.

Let $M\in\{0,1\}^D$ be the binary editing mask in latent space and decompose any noise as:
\begin{equation}
\epsilon^{\mathrm{in}}=M\odot \epsilon,\qquad
\epsilon^{\mathrm{base}}=(1-M)\odot \epsilon.
\label{eq:mask_decompose_main}
\end{equation}

For each candidate $\epsilon^{(i)}$, we apply the same decomposition and denote the resulting components as $\epsilon^{(i),\mathrm{in}}$ and $\epsilon^{(i),\mathrm{base}}$.
Neighbor GRPO forms candidates $\{\epsilon^{(i)}\}_{i=1}^G$ around an anchor $\epsilon^\ast$ using
global perturbation, so both $\epsilon^{(i),\mathrm{in}}$ and $\epsilon^{(i),\mathrm{base}}$
vary across candidates.
We decompose the terminal reward $R(x_0;c)$ in Eq.~\eqref{eq:grpo_adv} into an additive form,
$R(x_0;c)=R_{\mathrm{edit}}(x_0;c)+R_{\mathrm{pres}}(x_0;c)$, where $R_{\mathrm{edit}}$ measures
instruction adherence in the editing region and $R_{\mathrm{pres}}$ measures
non-target preservation in the background region.
Locally around the anchor (formalized in Appendix~A),
the preservation term is weakly sensitive to perturbations inside the editing region, yielding the approximation:
\begin{equation}
\begin{aligned}
R(x_0^{(i)};c)&\approx R_{\mathrm{edit}}(x_0^{(i)};c)\\
&\;+ R_{\mathrm{pres}}\!\Big(\Phi_\theta(\epsilon^{\mathrm{in},*}+\epsilon^{(i),\mathrm{base}};c);c\Big)
+ \Delta_{\mathrm{int}},
\end{aligned}
\label{eq:reward_decomp_main}
\end{equation}
where $\epsilon^\ast$ denotes the anchor initial noise, $\epsilon^{\mathrm{in},*}=M\odot\epsilon^\ast$ is its in-region component, and $\Delta_{\mathrm{int}}$ collects higher-order cross-region couplings and approximation residuals~\cite{hertz2023prompttoprompt,chefer2023attend}.

Under global perturbation, $\epsilon^{(i),\mathrm{base}}$ varies across candidates, causing the background preservation term in Eq.~\eqref{eq:reward_decomp_main} to fluctuate within each group. Because GRPO standardizes rewards to form group-normalized advantages (Eq.~\eqref{eq:grpo_adv}), this nuisance variance inflates within-group dispersion and lowers the effective SNR of the standardized advantages, weakening localized credit assignment. A first-order analysis (Appendix~A) shows that the resulting background-induced nuisance variance scales on the order of $\sigma^2$, whereas suppressing background perturbations with RDP reduces it to the order of $\alpha_{\mathrm{base}}^2$, thereby improving the SNR of group-standardized advantages. This motivates Region-Decoupled Perturbation (Sec.~\ref{sec:rdp}) to suppress background randomization, while residual cross-region couplings are captured by $\Delta_{\mathrm{int}}$ and addressed by ACD in Sec.~\ref{sec:acd}.

\section{Methodology}
\label{sec:method}

Our method improves GRPO credit assignment for deterministic ODE image editing by combining region-constrained exploration (RDP) and a trajectory-level intrinsic signal (ACD).
Both components, together with the editing mask $M$ they rely on, are used only during post-training; at inference the model is a standard fine-tuned editor that runs the default single-path deterministic ODE sampler without a mask and with no additional overhead.
We introduce RDP in Sec.~\ref{sec:rdp} and ACD in Sec.~\ref{sec:acd}, and present the overall training procedure in Figure~\ref{fig2}.

\begin{figure*}[t]
    \centering
    \includegraphics[width=0.98\textwidth,keepaspectratio]{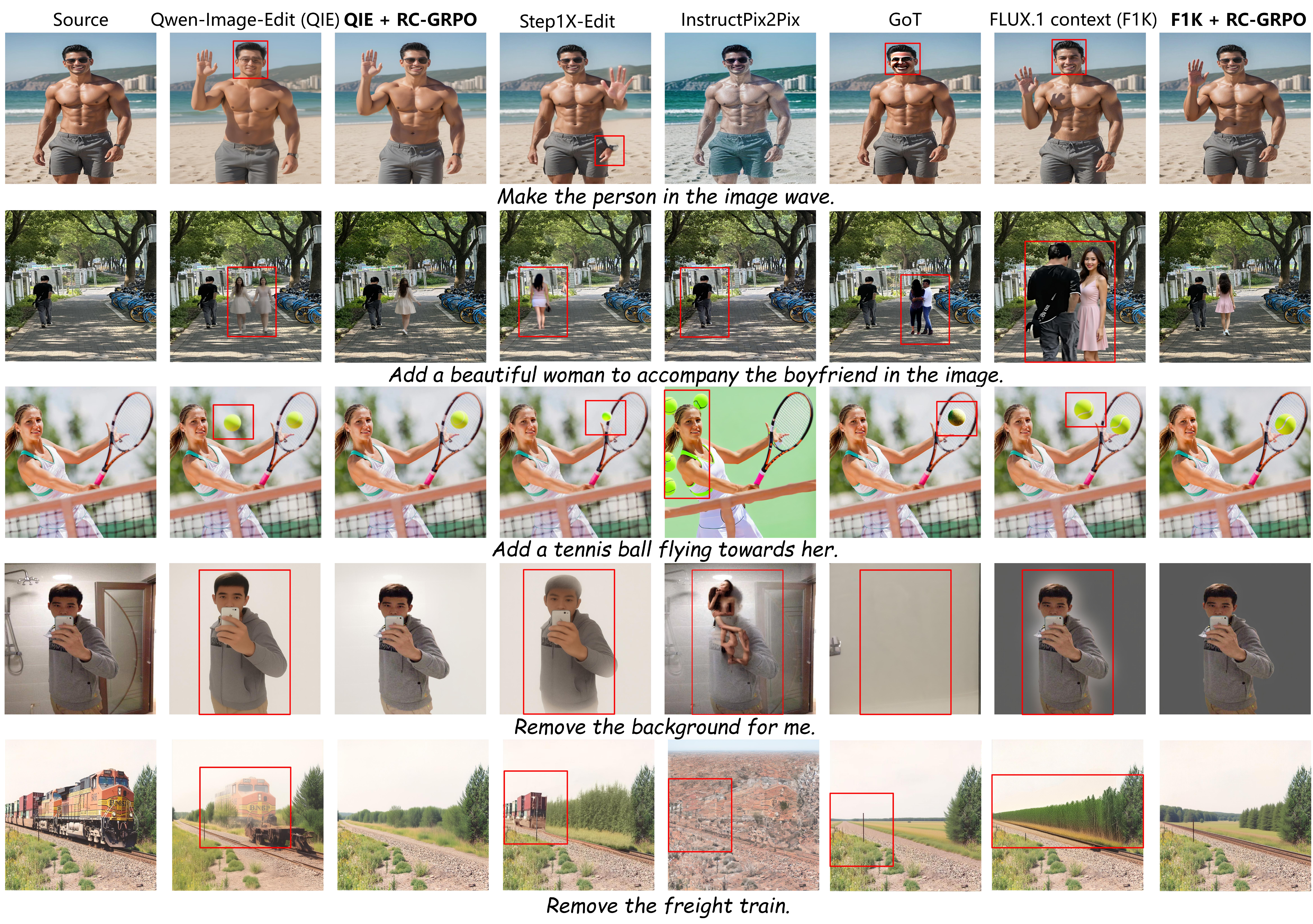}
    \vspace{-1mm}
    \caption{Visual comparisons for instruction-guided image editing.}
    \label{fig:qualitative}
    \vspace{-0.8mm}
\end{figure*}

\subsection{Region-Decoupled Perturbation}
\label{sec:rdp}

Our framework retains the surrogate leaping-policy training scheme of Neighbor GRPO~\cite{he2025neighbor}, which enables policy-gradient updates under deterministic ODE sampling. Our contribution is orthogonal to this scheme: we redesign how the candidate neighborhood is constructed, using region-decoupled rather than global perturbation. For consistency with this anisotropic construction, we also compare candidates under a mask-aware region-normalized distance in place of an isotropic Euclidean one.

Motivated by the variance-reduction analysis (Appendix~A), we construct a region-decoupled initial noise neighborhood using the editing mask $M$.
We perturb the initial noise primarily inside the editing region and suppress perturbations in the background.
This injects exploration only at $t{=}1$ while keeping subsequent rollouts fully deterministic ODE integrations under the same solver.
To stabilize candidate comparisons, we additionally include one unperturbed reference candidate in each group.
During GRPO updates, we perform symmetric anchor sampling only over the perturbed candidates, so that anchor selection remains exchangeable within this subset.

\subsubsection{Mask-Structured Initial Perturbation.}
To balance exploration in the editing region with pixel-level consistency in the background, we construct a spatially anisotropic perturbation coefficient map $\boldsymbol{\alpha}\in[0,1]^D$ aligned with $M$:
\begin{equation}
\boldsymbol{\alpha}
=
\alpha_{\mathrm{edit}} M + \alpha_{\mathrm{base}} (1-M).
\label{eq:alpha_map}
\end{equation}

We assign a larger coefficient $\alpha_{\mathrm{edit}}$ to promote exploration within the editing region.
For numerical stability, we set $\alpha_{\mathrm{base}}=\varepsilon_{\mathrm{bg}}$ with a small $\varepsilon_{\mathrm{bg}}$ instead of enforcing $\alpha_{\mathrm{base}}=0$, which approximately freezes background perturbations within each group.
Based on $\boldsymbol{\alpha}$, we sample a shared reference noise $\epsilon^\ast\sim\mathcal N(0,I)$ and construct a group of $G$ candidate initial noises.
We reserve one index $i_0\in\{1,\dots,G\}$ as the reference candidate and set it to be unperturbed.
The remaining $G{-}1$ candidates are sampled by anisotropic mixing:
\begin{equation}
\epsilon^{(i_0)}=\epsilon^\ast,\qquad
\epsilon^{(i)}=\sqrt{1-\boldsymbol{\alpha}^2}\odot \epsilon^\ast+\boldsymbol{\alpha}\odot\delta^{(i)},\ i\neq i_0,
\label{eq:masked_init_with_anchor}
\end{equation}
where $\odot$ and the square root are applied element-wise, and
$\delta^{(i)}\sim\mathcal N(0,I)$ are sampled independently for $i\neq i_0$.

This variance-preserving mixing keeps every candidate marginally $\mathcal N(0,I)$, consistent with the pre-trained editor's noise distribution at $t{=}1$.
Since the perturbations are spatially anisotropic, we compare candidates under a mask-structured, region-normalized distance (Appendix~B, Scheme~B).

\subsubsection{Deterministic Rollouts and Masked Surrogate Policy.}
Since per-step ODE transition densities are degenerate, we adopt the surrogate leaping policy of Neighbor GRPO as our training scaffold, but instantiate it over our region-decoupled candidates and with a mask-aware distance (Eq.~\eqref{eq:policy_over_G}).

We first roll out all candidates with a fixed deterministic integrator under $\theta_{\mathrm{old}}$:
\begin{equation}
x_t^{(i)} \triangleq \Phi_{1\rightarrow t}(\epsilon^{(i)};\theta_{\mathrm{old}},c), \qquad i=1,\dots,G.
\label{eq:candidates}
\end{equation}

Let $\mathcal I \triangleq \{1,\dots,G\}\setminus\{i_0\}$ denote the set of perturbed candidate indices.
For efficiency, we use anchor subsampling: for each group, we sample a set of anchors
$\mathcal K \subseteq \mathcal I$ with $|\mathcal K|=B$ uniformly without replacement.
For each anchor $k\in\mathcal K$, we define the anchor trajectory under current parameters:
\begin{equation}
\bar{x}_t^{(k)} \triangleq \Phi_{1\rightarrow t}(\epsilon^{(k)};\theta,c).
\label{eq:current_anchor}
\end{equation}

We then instantiate the surrogate policy as a softmax over a mask-aware distance $d_C^2(\cdot,\cdot;M)$.
Because the editing region $M$ and the background $(1-M)$ can be highly imbalanced in size, a naive Euclidean distance is often dominated by background deviations. To address this, we use a region-normalized masked metric (detailed in Appendix~B, Eq.~(B.3)), where $\tau_{\mathrm{edit}}$ and $\tau_{\mathrm{base}}$ are region-wise normalization hyperparameters (values in Appendix~C):
\begin{equation}
\pi_\theta\!\left(i \mid t,k\right)
=
\frac{\exp\!\left(-\frac{1}{2}\,d_C^2\!\left(x_t^{(i)},\bar{x}_t^{(k)};M\right)\right)}
{\sum_{j=1}^G\exp\!\left(-\frac{1}{2}\,d_C^2\!\left(x_t^{(j)},\bar{x}_t^{(k)};M\right)\right)} .
\label{eq:policy_over_G}
\end{equation}

Using the same form with the old anchor trajectory $x_t^{(k)}$ from Eq.~\eqref{eq:candidates} defines $\pi_{\theta_{\mathrm{old}}}(i\mid t,k)$, and we compute the per-step importance ratio as:
\begin{equation}
\rho_t^{(i\mid k)}(\theta)
=
\frac{
\pi_\theta(i \mid t,k)
}{
\pi_{\theta_{\mathrm{old}}}(i \mid t,k)
}.
\label{eq:ratio}
\end{equation}

We use $\rho_t^{(i\mid k)}(\theta)$ as the per-step ratio in the standard clipped GRPO objective.
In practice, we evaluate the GRPO objective for each anchor $k\in\mathcal K$ and average over anchors.

\subsubsection{Training-Time Surrogate and Inference Usage.}
The surrogate is a differentiable, mask-aware parameterization of relative preference over the candidate set rolled out under $\theta_{\mathrm{old}}$, not an exact density ratio; the unperturbed reference $i_0$ participates as a comparison item while anchors are sampled symmetrically from the perturbed subset $\mathcal I$. At inference we set $\alpha_{\mathrm{edit}}=\alpha_{\mathrm{base}}=0$, so no candidate neighborhood or surrogate policy is constructed and sampling reduces to the default single deterministic path.

\subsection{Process Alignment via Attention Concentration Density}
\label{sec:acd}

Beyond the variance-reduction view in Sec.~\ref{sec:motivation}, Region-Decoupled Perturbation (Sec.~\ref{sec:rdp}) localizes exploration only at initialization ($t{=}1$) and does not guarantee edit locality throughout the rollout. In practice, global attention interactions in Transformers can cause edit-related signals to propagate beyond the intended editing region, leading to excessive cross-attention spillover into non-target regions.
This excessive spillover increases the residual cross-region term $\Delta_{\mathrm{int}}$ in Eq.~\eqref{eq:reward_decomp_main} and can weaken localized credit assignment. We therefore introduce Attention Concentration Density (ACD) as a trajectory-level intrinsic reward to encourage region-focused attention over the entire rollout.

\subsubsection{Attention Concentration Density (ACD).}
Let $A^{(i)}_{l,t}(v,u)$ denote the text-to-image cross-attention weight (aggregated over heads) at layer $l$ and sampling step $t$,
where text token $v$ attends to image token $u$ (thus $\sum_u A^{(i)}_{l,t}(v,u)=1$ for each $v$).
We define the attention mass received by each image token as:
\begin{equation}
a^{(i)}_{l,t}(u) = \sum_{v} A^{(i)}_{l,t}(v,u).
\label{eq:acd_mass}
\end{equation}

We then measure the \emph{relative attention density} inside the editing mask $M$ by comparing the average mass inside $M$
to the global average over all image tokens:
\begin{equation}
\mathrm{ACD}^{(i)}_{l,t}
=
\frac{
\frac{1}{|M|+\varepsilon}\sum_{u} a^{(i)}_{l,t}(u)\, M(u)
}{
\frac{1}{N_{\mathrm{img}}+\varepsilon}\sum_{u} a^{(i)}_{l,t}(u)
},
\label{eq:acd_def}
\end{equation}
where $|M|=\sum_u M(u)$, $N_{\mathrm{img}}$ is the number of image tokens, and $\varepsilon$ is a small constant.
Here $\mathrm{ACD}^{(i)}_{l,t}>1$ means that attention density inside the editing region exceeds the global average.

\subsubsection{Intrinsic Reward Aggregation.}
We aggregate ACD over a predefined set of late cross-attention layers $\mathcal{L}_{\mathrm{sel}}$ and all sampling steps to form an intrinsic reward:
\begin{equation}
R_{\mathrm{acd}}^{(i)}=\frac{1}{|\mathcal{L}_{\mathrm{sel}}|\,T}\sum_{l\in\mathcal{L}_{\mathrm{sel}}}\sum_{t=1}^{T}\mathrm{ACD}^{(i)}_{l,t}.
\label{eq:acd_reward}
\end{equation}

We set $\mathcal{L}_{\mathrm{sel}}$ to the last $K$ cross-attention layers, which tend to give more consistent separation between successful and failed rollouts; the value of $K$ per backbone is specified in Appendix~C (diagnostic in Appendix~D.1).
Along the temporal axis, we aggregate ACD uniformly over all $T$ sampling steps, treating each denoising step equally rather than selecting or reweighting particular steps.
During training, we add $R_{\mathrm{acd}}$ to the task reward to reduce excessive cross-attention spillover into non-target regions.

\section{Experiments}
\label{sec:experiments}

\subsection{Experimental Settings}
\label{sec:exp_setup}

\subsubsection{Dataset.}
For post-training, we use a curated six-category subset of ImgEdit~\cite{ye2025imgedit}.
We evaluate on the Add, Remove, and Replace subsets of CompBench~\cite{jia2026compbench}, containing 982, 1,331, and 152 examples, respectively (2,465 total).
Each example provides a source image, an instruction, and an edit mask; we use the official evaluation scripts.
We also evaluate on held-out GEdit-Bench~\cite{liu2025step1x} (606 English samples, 11 task types) and ImgEdit-Bench~\cite{ye2025imgedit} (Basic split, 737 samples, 9 task types).

\subsubsection{Evaluation Metrics.}
Following CompBench, we measure non-target preservation with background PSNR, SSIM~\cite{wang2004image}, and LPIPS~\cite{zhang2018unreasonable}.
For editing-region adherence, LC-I is the CLIP~\cite{radford2021learning} similarity between edited and ground-truth foreground crops, while LC-T compares the edited foreground with its target description.
GEdit-Bench reports $Q_O=\sqrt{Q_{\mathrm{SC}}Q_{\mathrm{PQ}}}$ from semantic consistency and perceptual quality scores ($0$--$10$), whereas ImgEdit-Bench uses ImgEdit-Judge ($1$--$5$); Appendix~E provides full definitions.

\subsubsection{Implementation Details.}
We train attention-projection LoRA~\cite{hu2022lora} adapters with GRPO using an EditScore~\cite{luo2026editscore} task reward and $R_{\mathrm{acd}}$, each normalized across the minibatch before computing group-standardized advantages.
Appendix~C specifies all training hyperparameters and matched evaluation configurations.

\subsection{Main Results}
\label{sec:main_results}

\subsubsection{Quantitative Comparison.}
Table~\ref{tab:compbench_main} compares RC-GRPO-Editing with representative published editors, including Bagel~\cite{deng2025bagel}, and its corresponding base models on CompBench.

\begin{table}[t]
\centering
\footnotesize
\setlength{\tabcolsep}{1pt}
\renewcommand{\arraystretch}{1.05}
\begin{tabular}{@{}>{\raggedright\arraybackslash}p{0.33\columnwidth}ccccc@{}}
\toprule
\textbf{Method} & \textbf{LC-I}~$\uparrow$ & \textbf{LC-T}~$\uparrow$ & \textbf{PSNR}~$\uparrow$ & \textbf{SSIM}~$\uparrow$ & \textbf{LPIPS}~$\downarrow$ \\
\midrule
InstructPix2Pix & 0.777 & 19.445 & 21.416 & 0.695 & 0.137 \\
Step1X-Edit & 0.817 & 20.501 & 23.371 & 0.882 & 0.078 \\
GoT & 0.807 & 20.268 & 24.675 & 0.890 & \underline{0.067} \\
Bagel & 0.827 & 20.864 & \underline{25.389} & 0.891 & \underline{0.067} \\
FLUX.1-Kontext & 0.818 & 21.027 & 24.585 & 0.890 & 0.071 \\
\quad + RC-GRPO (ours) & \underline{0.830} & \textbf{21.588} & \textbf{25.532} & \underline{0.927} & \textbf{0.052} \\
Qwen-Image-Edit & 0.819 & 21.332 & 24.394 & 0.772 & 0.112 \\
\quad + RC-GRPO (ours) & \textbf{0.834} & \underline{21.520} & 25.385 & \textbf{0.973} & 0.086 \\
\bottomrule
\end{tabular}
\caption{CompBench foreground correctness and background preservation. Best in \textbf{bold}, second best \underline{underlined}.}
\label{tab:compbench_main}
\end{table}

Under matched base-versus-post-training comparisons, RC-GRPO improves all five CompBench metrics on both editing backbones (Table~\ref{tab:compbench_main}). On FLUX.1-Kontext, it raises LC-I from $0.818$ to $0.830$, LC-T from $21.027$ to $21.588$, PSNR from $24.585$ to $25.532$, and SSIM from $0.890$ to $0.927$, while reducing LPIPS from $0.071$ to $0.052$. On Qwen-Image-Edit, the corresponding values change from $0.819$ to $0.834$, $21.332$ to $21.520$, $24.394$ to $25.385$, $0.772$ to $0.973$, and $0.112$ to $0.086$.

\begin{table}[t]
\centering
\small
\setlength{\tabcolsep}{3pt}
\renewcommand{\arraystretch}{1.05}
\begin{tabular*}{\columnwidth}{@{\extracolsep{\fill}}lccc@{}}
\toprule
\textbf{Method} & $Q_{\mathrm{SC}}$$\uparrow$ & $Q_{\mathrm{PQ}}$$\uparrow$ & $Q_O$$\uparrow$ \\
\midrule
Gemini 2 Flash$^*$ & 7.274 & 7.327 & 6.971 \\
Doubao$^*$ & 7.353 & \underline{7.651} & 7.230 \\
GPT-4o$^*$ & \underline{7.847} & \textbf{7.705} & \textbf{7.692} \\
\midrule
Instruct-Pix2Pix & 4.746 & 6.913 & 4.578 \\
AnyEdit & 3.713 & 6.730 & 3.635 \\
MagicBrush & 5.752 & 7.069 & 5.558 \\
OmniGen & 6.845 & 6.700 & 6.352 \\
Step1X-Edit & 7.388 & 7.279 & 7.067 \\
Step1X-Edit v1.1 & 7.652 & 7.408 & 7.346 \\
FLUX.1-Kontext-dev & 6.570 & 7.218 & 6.282 \\
\quad + RC-GRPO (ours) & 6.583~{\scriptsize(+0.013)} & 7.276~{\scriptsize(+0.058)} & 6.347~{\scriptsize(+0.065)} \\
Qwen-Image-Edit & 7.769 & 7.535 & 7.494 \\
\quad + RC-GRPO (ours) & \textbf{7.943}~{\scriptsize(+0.174)} & 7.602~{\scriptsize(+0.067)} & \underline{7.674}~{\scriptsize(+0.180)} \\
\bottomrule
\end{tabular*}
\caption{Comparison with published editors on GEdit-Bench (EN, full set, $0$--$10$). $^*$ denotes closed-source methods. All scores use the official Qwen2.5-VL-72B judge; baseline numbers are taken from the GEdit-Bench release~\cite{liu2025step1x}. Best in \textbf{bold}, second best \underline{underlined}.}
\label{tab:gedit_sota}
\end{table}

On GEdit-Bench, RC-GRPO with Qwen-Image-Edit attains the highest $Q_{\mathrm{SC}}$ and second-highest $Q_O$ among baselines including AnyEdit~\cite{yu2025anyedit} and OmniGen~\cite{xiao2025omnigen} (Table~\ref{tab:gedit_sota}). Its $Q_O$ of $7.674$ surpasses all other open-source methods as well as the closed-source Gemini 2 Flash and Doubao, and is within $0.018$ of GPT-4o.

\subsubsection{Qualitative Comparison.}
Figure~\ref{fig:qualitative} presents representative comparisons.
Competing editors often suffer from \emph{under-editing} (partial target realization) or \emph{over-editing} (spillover artifacts), whereas our framework mitigates both, yielding edits that are semantically consistent inside the mask and structurally stable outside.

\begin{figure*}[t]
\centering
\includegraphics[width=\textwidth,keepaspectratio]{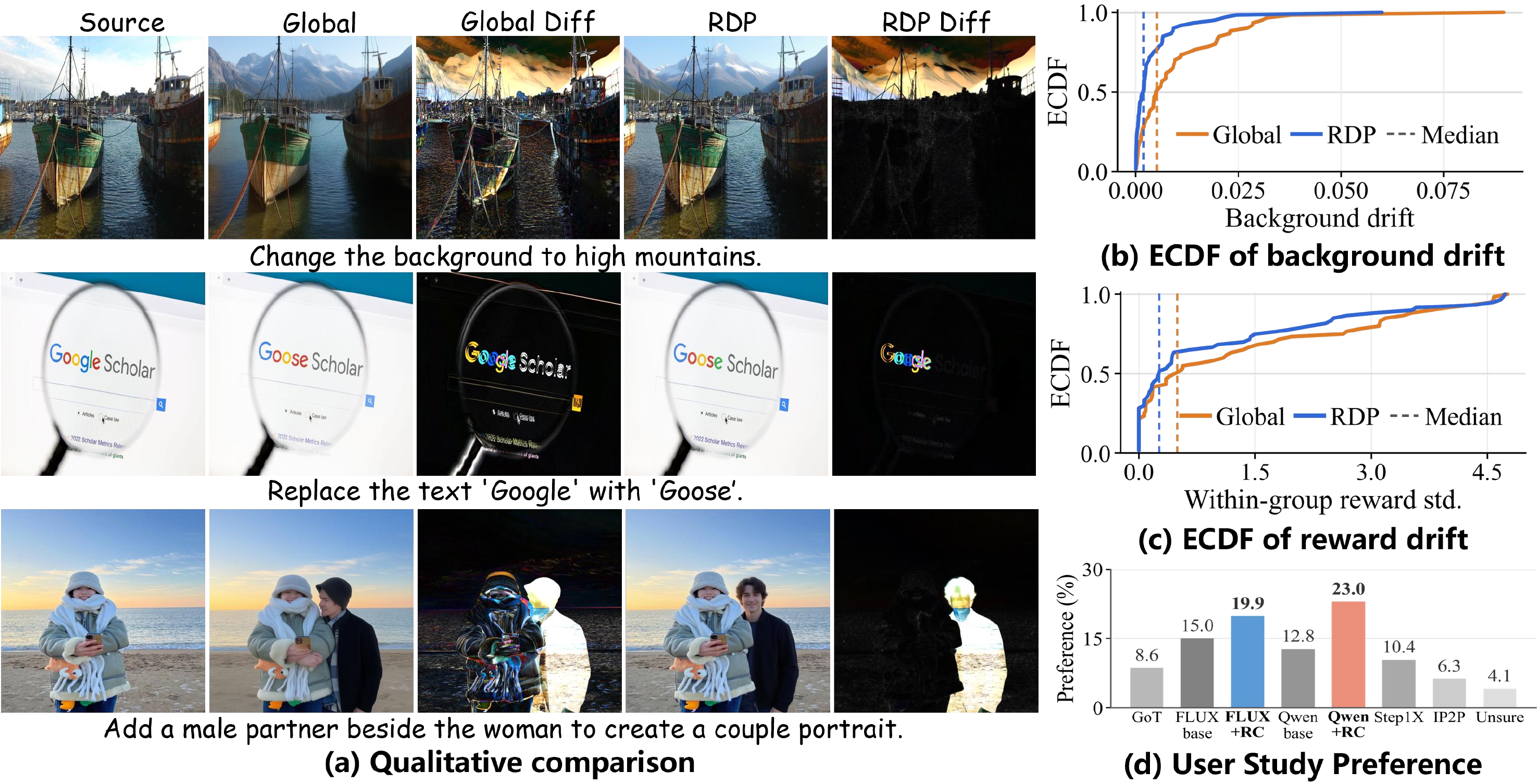}
\caption{\textbf{Analysis of region-decoupled exploration and user preferences.}
(a) Qualitative examples for global exploration and RDP. ``Diff'' denotes the visual difference between each edited output and its source image, with brighter regions indicating larger changes.
(b) ECDF of background drift on the non-target region. (c) ECDF of within-group reward dispersion, measured by reward std.
Dashed lines denote medians. (d) User-study preference rates.}
\label{fig:qual_gp_rdp}
\end{figure*}

\subsubsection{User Study.}
\label{sec:user_study}
Across 800 trials, participants viewed the source image, instruction, and anonymized outputs in randomized order, then selected the best result or chose Not sure.
Figure~\ref{fig:qual_gp_rdp}(d) shows that Qwen-Image-Edit + RC-GRPO received the highest preference rate at $23.0\%$ (184/800), followed by FLUX.1-Kontext-dev + RC-GRPO at $19.9\%$ (159/800), exceeding their corresponding base rates of $12.8\%$ (102/800) and $15.0\%$ (120/800), respectively.

\subsection{Ablation Studies}
\label{sec:ablation}

We isolate RDP and ACD under the same backbone, LoRA parameterization, reward, GRPO budget, and inference settings, varying only the $t{=}1$ exploration scheme and the use of ACD (Table~\ref{tab:ablation_main}).
The global variant adapts Neighbor GRPO's initial-noise perturbation~\cite{he2025neighbor} to FLUX.1-Kontext.

\paragraph{Effect of Region-Decoupled Exploration.}
With ACD disabled, Figure~\ref{fig:qual_gp_rdp}(a--c) shows that RDP confines changes to the intended region, reduces non-target background drift, and tightens within-group reward std relative to global exploration, supporting more stable group-wise credit assignment.

\paragraph{Contributions of RDP and ACD.}
Table~\ref{tab:ablation_main} shows that global GRPO raises LC-T from $21.027$ to $21.560$ but degrades all three background metrics, whereas RDP improves all five metrics relative to global exploration.
Adding ACD partly restores background fidelity under global exploration and achieves the best result on all five metrics when combined with RDP.
The two components act at complementary stages: RDP suppresses nuisance variance at initialization, while ACD reduces attention spillover along the trajectory.

\begin{table}[t]
\centering
\small
\setlength{\tabcolsep}{2pt}
\renewcommand{\arraystretch}{1.05}
\begin{tabular*}{\columnwidth}{@{\extracolsep{\fill}}lccccc@{}}
\toprule
Variant & LC-I$\uparrow$ & LC-T$\uparrow$ & PSNR$\uparrow$ & SSIM$\uparrow$ & LPIPS$\downarrow$ \\
\midrule
Base & 0.818 & 21.027 & 24.585 & 0.890 & 0.071 \\
GRPO (global) & 0.823 & 21.560 & 24.380 & 0.884 & 0.075 \\
GRPO (global) + ACD & 0.823 & 21.548 & 24.880 & 0.905 & 0.064 \\
GRPO + RDP & 0.827 & 21.572 & 24.950 & 0.902 & 0.060 \\
RDP + ACD (ours) & \textbf{0.830} & \textbf{21.588} & \textbf{25.532} & \textbf{0.927} & \textbf{0.052} \\
\bottomrule
\end{tabular*}
\caption{Effect of core components (RDP and ACD) under a fixed post-training budget.}
\label{tab:ablation_main}
\end{table}

\subsubsection{Generalization Across Benchmarks and Backbones.}
On ImgEdit-Bench, RC-GRPO consistently improves both FLUX.1-Kontext and Qwen-Image-Edit, and the gains cover all nine edit types on each backbone (Appendix~E, Table~6).
Although RDP localizes exploration to the editing region, its gains also extend to globally oriented edits: Background and Tone improve on both backbones in GEdit-Bench (Appendix~E, Table~3), and Background editing improves on both backbones in ImgEdit-Bench (Appendix~E, Table~6). These results support the generality of our approach across architectures.
For ImgEdit-Bench, we use ImgEdit-Judge, the open-source evaluator officially released with the benchmark, rather than the GPT-4o judge used by its official leaderboard; we therefore restrict the comparison to the controlled base-versus-post-trained setting in Table~\ref{tab:cross_benchmark}.

\begin{table}[t]
\centering
\small
\setlength{\tabcolsep}{3pt}
\renewcommand{\arraystretch}{1.05}
\begin{tabular*}{\columnwidth}{@{\extracolsep{\fill}}lccc@{}}
\toprule
\textbf{Backbone} & \textbf{Base} & \textbf{+ RC-GRPO (ours)} & \textbf{Gain}$\uparrow$ \\
\midrule
FLUX.1-Kontext & 4.302 & \textbf{4.411} & +0.109 \\
Qwen-Image-Edit & 4.402 & \textbf{4.482} & +0.080 \\
\bottomrule
\end{tabular*}
\caption{Cross-backbone generalization on ImgEdit-Bench (Basic, 737), evaluated by ImgEdit-Judge ($1$--$5$). Gain is measured over the corresponding base model.}
\label{tab:cross_benchmark}
\end{table}
\FloatBarrier

\subsubsection{Discussion.}
Our method improves localized credit assignment for deterministic ODE editing via region-constrained exploration (RDP) and a trajectory-level intrinsic signal (ACD). One limitation remains.
RDP enforces edit locality only at initialization ($t{=}1$); residual cross-region couplings can still affect non-target regions, especially for globally consistent edits such as lighting and reflections.
A natural direction for future work is to extend region-constrained exploration from initialization to the full denoising trajectory, so that locality is enforced across timesteps rather than only at $t{=}1$.

\section{Conclusion}
We present RC-GRPO-Editing, which resolves the credit assignment mismatch in flow-based image editing via Region-Decoupled Perturbation (RDP) and Attention Concentration Density (ACD). By suppressing background nuisance variance, it balances instruction adherence and non-target preservation, outperforming strong baselines on CompBench, GEdit, and ImgEdit and in a user study.

\bibliography{main}

\end{document}